\documentclass{article}
\usepackage{kr}

\usepackage{comment}
\usepackage{float}

\usepackage{times}
\usepackage{soul}
\usepackage{url}
\usepackage[hidelinks]{hyperref}
\usepackage[utf8]{inputenc}
\usepackage[small]{caption}
\usepackage{graphicx}
\usepackage{amsmath}
\usepackage{amsthm}
\usepackage{booktabs}
\usepackage{algorithm}
\usepackage{algorithmic}
\usepackage{multirow}

\title{Specific Domain Ontology Construction Using Large Language Models}

\author{%
Vivian Magri Alcadi Soares$^{1,2}$\and
Renata Wassermann$^{1,2}$ \\
\affiliations
$^1$University of São Paulo (USP)\\
$^2$Center for Artificial Intelligence (C4AI)\\
\emails
\{vivian.soares@alumni.usp.br, renata@ime.usp.br\}
}

\begin{document}

\maketitle

\begin{abstract}
    Ontologies are useful structures to organize and maintain information that can be understood both by humans and systems. However, since their manual crafting is a laborious task, many specific domains lack reference ontologies. The outstanding ability for understanding natural language demonstrated by the Large Language Models (LLMs) has motivated their application to aid on a variety of fields, including on ontology development. This work presents the experimentation with a technique that uses LLMs in the role of domain experts to build conceptual hierarchies for a given initial concept. Twenty ontologies automatically constructed for the domain of the Brazilian maritime territory (a.k.a the Blue Amazon) using GPT-3.5 and GPT-4 were then evaluated by human experts. The models were able to construct overall coherent conceptualizations of the domain, but none of the outputs was completely satisfactory as a representation of the context without refinement.
\end{abstract}

\section{Introduction}

Developing an ontology defines a common vocabulary for researchers who need to share information of a domain, and is akin to building machine-interpretable definitions for a set of concepts and relations among them 
\cite{Onto_101}. The manual construction of ontologies, however, is a strenuous endeavor, that requires appropriate knowledge. Many (semi-)automatic ontology extraction methods have been proposed over the years, ranging from basic rule-based and statistical approaches to complicated Machine Learning (ML) and hybrid architectures. Nevertheless, the problem of obtaining a well-structured, relevant and coherent with the desirable application ontology without a fair amount of human labor persisted.

In this scenario, approaches using Large Language Models (LLMs) to develop ontologies have shown promising results. The methods, that seek to leverage the linguistic abilities these models gain on their extensive pre-training, 
appear as promising to bridge the gap of ontologies for specific domains that haven't yet been provided a gold standard. Such is the case of the Brazilian maritime territory, a vast area with approximately the same size as the Amazon rainforest, and also referred to as the "Blue Amazon". It is not only a region of great economic and commercial value for Brazil, but the importance of its resources and multiple ecosystems is invaluable, also playing a key role in climate regulation. Yet, the Blue Amazon is not well-known to the wider public. Information about it is dispersed in academic volumes and government reports, or in obscure databases \cite{2022_BLAB}. 

This research proposes an experiment with a LLM-enabled algorithm to aid on the organization of information about the specific domain of the Blue Amazon.
Since the application of methodologies of this kind to ontology learning is still recent and the evaluation process is not standardized, it was considered essential to also construct and conduct a thorough assessment of the outputs, involving humans with deep knowledge of the domain.
The analysis of the results reveals the usefulness that can be perceived on the employment of generative AI to accelerate the construction of ontologies. Nonetheless, it also indicates that the full automation of this process is not yet recommended.


\section{Literature Review}

Handcrafting big ontologies is a difficult task. Over the past decades, various methods using linguistic, statistical, and logical techniques, usually associated with steps of the process known as the Ontology Learning Layer Cake, have contributed to the improvement of automatic ontology development \cite{Srv2018}. More recently, we observed a great rise in the use of deep neural network-based procedures for Ontology Learning (OL) \cite{2023_SCRE,2025_Amalki-etal}. According to Reshadat
et al.~\shortcite{2023_SCRE}, the key advantage that makes Deep Learning (DL) such a powerful approach is that the feature engineering procedure is done automatically. Also, Du et al.~\shortcite{2024_Du-etal} mention how extensive works have demonstrated that DL-based approaches excelled in understanding texts compared to shallow learning.

However, deep learning techniques come with some drawbacks. Ontologies require precise representations, and these models show limited success for taxonomy induction \cite{2024_Du-etal}. They achieve good performance extracting information on the form of knowledge graphs (KGs), but lacking the formal, hierarchical organization of concepts of an ontology that enable further reasoning.
They also might struggle with capturing subtle semantic nuances or understanding contextual variations within domain-specific terminology \cite{2024_Du-etal}.

Du et al.~\shortcite{2024_Du-etal} also claims that adapting DL models to new domains and utilizing transfer learning techniques are challenging tasks. Usually these methods require an annotated corpus with concepts and the relations between them \cite{2023_SCRE}, and acquiring annotated ontologies or extensive labeled data for specific and complex domain knowledge for the training might hinder DL widespread application for OL, not to mention these models large demand for computational resources \cite{2024_Du-etal}. On top of that, building and fine-tuning deep learning models require specialized expertise, which can limit accessibility and practical implementation for non-experts when applying it to different domain ontologies \cite{2024_Du-etal}.

Amidst this landscape, the emergence of Large Language Models (LLMs) stands as a disruptive force \cite{2024_Du-etal}. The systematic mapping study of DL-driven OL by Amalki, Tatane and Bouzit~\shortcite{2025_Amalki-etal} show a spike in publications in 2023, coinciding with the emergence of advanced models such as transformers and LLMs, marking a peak in innovation. Du et al.~\shortcite{2024_Du-etal} claim these models exhibit a remarkable aptitude for understanding semantic nuances, capturing context and inferring relationships among entities. Leveraging the prowess of pre-trained language representations, few-shot learning techniques, and harnessing the inherent linguistic and conceptual understanding embedded within these models could mitigate the data dependency, domain adaptation and computational resources challenges, enabling ontology construction with smaller datasets and lower computational requirements \cite{2024_Du-etal}.

The most common OL tasks addressed in the works reviewed by Amalki, Tatane and Bouzit~\shortcite{2025_Amalki-etal} are construction, enrichment and population of ontologies, and the most frequent subtasks mapped are term extraction, relation extraction and ontology refinement. They view the prevalence of experimental and empirical approaches they found in the research analyzed as revealing of a primary concern in method development.
Du et al.~\shortcite{2024_Du-etal} also includes a summary of empirical attempts to verify if LLMs are suitable for OL. The emphasis of these experiments is on term typing and relation extraction, specially for hierarchy building. They claim several studies have indicated that the utilization of LLMs to facilitate the identification of taxonomy significantly mitigates the need for manual intervention.

Among such experiments, the work of Funk et al~\shortcite{2023_Onto_LLMs}, using OpenAI's GPT 3.5 API to fully automatically construct simple ontologies consisting of hierarchical conceptualizations, stands out for its simplicity of replication for a variety of domains.
Their prompts are mostly based on a mild form of few-shot learning that takes as input from the user a seed concept \textit{C0}. The \textit{C0} is then inserted as the root of \textit{H}, the hierarchy being constructed, and goes through an exploration loop where the LLM provides a textual description for the concept under analysis and attempts to identify subconcepts for it. The listed subconcepts that are judged as relevant by the model are placed into \textit{H}, and the execution continues until either the algorithm finishes exploring all the concepts that were included, or the stop conditions defined by the hyperparameters are met.

The work of Perera and Liu~\shortcite{2024_Perera-Liu} highlights the dynamic and progressive nature of Generative AI technology, concluding that LLMs continue to prove to be more efficient and scalable compared to traditional ML and manual methods. Fully automated LLM-based approaches, however, they consider feasible but challenging, necessitating human oversight at the current state of Generative AI. The view on the matter shared by Du et al.~\shortcite{2024_Du-etal} is that  completely automatic construction for ontologies by a system is appealing, but not likely to be possible. Their proposal is to investigate the utilization of interactive methodologies that involve domain experts in the knowledge acquisition process as a solution to improve the interpretive abilities of LLMs, as opposed to solely depending on prompting engineering.
Many works in the reviews of Perera and Liu~\shortcite{2024_Perera-Liu} and Du et al.~\shortcite{2024_Du-etal} indeed mention human feedback.
Perera and Liu~\shortcite{2024_Perera-Liu} claim these debates reflect the ongoing efforts to strike a balance between automation and accuracy. They defend that, by combining computational efficiency with human insights, we ensure the developed ontologies are not only technically sound but also practically meaningful.

Research also reveals the lack of standardization in the areas of evaluation metrics and benchmarks for ontology development \cite{2025_Amalki-etal,2024_Du-etal}. Perera and Liu~\shortcite{2024_Perera-Liu} report recent developments in this field have seen the integration of detailed evaluation metrics and techniques to further enhance the quality and applicability of the generated ontologies. Nonetheless, we observe that ontology evaluation continues to be a problem worthy of study \cite{2024_Du-etal}. Precision, recall, and F1 score are typical metrics for performance assessment in the field of ML, and many experiments with LLMs for ontology development still employ them to evaluate a model's ability to correctly identify relevant components while minimizing the inclusion of irrelevant or incorrect content \cite{2024_Perera-Liu}. Amalki, Tatane and Bouzit~\shortcite{2025_Amalki-etal}, however, advert that, although these metrics are important, their central role in evaluating a range of OL tasks probably points to serious limitations of related research work.
Different from a single information extraction task like relation or event extraction, the procedure to construct an ontology contains other elements and steps, and properly evaluating them is complex \cite{2024_Du-etal}.
Some tasks require specialized evaluation measures to capture the distinct aspects of their processes, suggesting the need to balance standardization and adaptability \cite{2025_Amalki-etal}.

Amalki, Tatane and Bouzit~\shortcite{2025_Amalki-etal} observe a sharp division in the literature. Whereas some fields rely on established benchmarks, like biomedical and bioinformatics domains, others require further development for evaluation protocols to be consistently effective. 
In fact, they highlight the predominance of biological and life sciences in the application of OL. Together, the domains of bioinformatics, biomedical, and healthcare represent 35.4\% of the studies. They argue, however, that the category of other domains (9 domains), with a total of 29.2\%, demonstrates not only the breadth but also the versatility of these techniques for ontology learning in capturing the diverse range of niche and emerging fields.
Both Amalki, Tatane and Bouzit~\shortcite{2025_Amalki-etal} and Du et al.~\shortcite{2024_Du-etal} recommend future research focusing on creating more extensive and field-specific benchmarks that can effectively measure the accuracy, relevance, completeness, and practical utility of ontologies generated by LLMs for a broader range of domains.

\section{Methodology}

The development of this project involved two main parts, the generation of ontologies related to the context of the Brazilian Ocean based on the algorithm described in the work of Funk et al~\shortcite{2023_Onto_LLMs} and made available by the authors\footnote{https://git.informatik.uni-leipzig.de/hosemann/onto-llm}, and the construction and conduction of the evaluation of the selected ontologies. The full outputs of the executions analyzed in this research, as well as the code versions and models of the questionnaires created, may be seen on the Ontology-for-Blue-Amazon repository\footnote{https://github.com/Vivian-Magri/Ontology-for-Blue-Amazon/}.

\subsection{Ontology Generation}

The base algorithm used for our ontology generation can be customized by a series of hyperparameter. Generally speaking, they define the limits for each execution, the caching of generated information and the space of probability for the tokens in the response. Two of those parameters were considered of particular interest, Exploration depth, that defines up to what point concepts will be explored, and Frequency threshold, which determines the minimum a subconcept must appear on the algorithm's repetitive listing to be considered.
The first stage of the experimentation was a grid search to determine which values for them seemed more suitable for the generation of ontologies on the desired domain.
Using OpenAI’s GPT API and the model 3.5 turbo — the default model operating by the time of the experiments (March 2024), and also the model mainly used on the researchers' experiments while designing the algorithm —, we tested the following combination of values (in accordance with the authors' recommendations of range for tests):
\begin{itemize}
    \item Exploration depth: 2 and 3
    \item Frequency threshold: from 5 to 20, with a step of 5
\end{itemize}

The other hyperparameters were kept on their default value, with the exception of the Prompt caching, that was deactivated, matching Sample caching — deactivated by default —, to avoid as much as possible interferences between executions through previously stored information. 
As initial concept (\textit{C0}), we first attempted to input "Blue Amazon" directly, but the executions of the algorithm finished without obtaining any verified sub-concepts for it. As a workaround, "Brazilian Water Resources", a more generally named concept, but still related to the original theme, was chosen.

The analysis of the ontologies produced by this first experiment showed that, as expected, lower frequency thresholds allow for a broader expansion of concepts, which makes them useful for exploration. With a frequency as low as 5, however, the expansion might go too far, and the inclusion of sub-concepts that do not relate well with the context increases. On the other hand, from the frequency threshold of 15 on, we observe a quite restricted expansion, which makes the resulting conceptualizations seem incomplete. Using an exploration depth of 2 has that same effect.
We then concluded that the combination of Exploration Depth = 3 and Frequency Threshold = 10 produced the most interesting result for our domain.
Therefore, these were the chosen values to apply on the subsequent experiments.

For the second and main stage of our ontologies generation, the chosen initial concept was "Coastal ecosystems". We looked for a \textit{C0} that would be specific enough to avoid overly generic ontologies, while not being too limited to allow for interesting expansions. Also, it is a concept defining a subdomain well under the Blue Amazon domain, but that could occur in different contexts, which enabled the evaluation of this aspect as well.

Considering what was observed on the first tests 
on how the results would change drastically due to small variations on how the \textit{C0} was input, we decided to test the chosen concept in different ways to compare the results. The aspects tested in combination were:
\begin{itemize}
    \item first letter of the central concept capitalized or not (keeping the rest in lowercase in both cases, like the concepts normally outputted by the algorithm); 
    \item explicitly referencing Brazil or not;
    \item \textit{C0} in English or in Brazilian Portuguese (PT-BR);
    \item requesting outputs to be in Brazilian Portuguese or not specified (only for the concepts in PT-BR).
\end{itemize}

Akin to the choice of including or not a direct reference to Brazil on the input, the variation on the language was introduced to investigate whether using the country's official language would affect the context of the outputted ontologies.
At the same time, we thought it would be interesting to see the effect of testing a language the active GPT models already exhibited good ability to communicate with, but which is not the one they were mainly trained in.

For a similar reason, we decided to assess the effect of inhibiting language mixing we observed for \textit{C0}s in PT-BR by creating a version of the code — that we will refer to as PT-out code version — where we added an instruction to the original prompts related to the listing, naming or description of concepts requesting the answer to be delivered in Brazilian Portuguese.
Another small edition to the code was made to facilitate running the algorithm using other GPT models. This allowed for a practical way to replicate the experiments in both GPT-3.5 and GPT-4 — the larger and most recent model of the family at the time — to compare the perceived quality of the results.

\begin{table*}[t] 
    \resizebox{\textwidth}{!}{
        \begin{tabular}{llrrl}
        \toprule
         &  & Number of Concepts & Subsumptions & Date of Execution \\
        Model & C0 &  &  &  \\
        \midrule
        \multirow[t]{13}{*}{3.5 Turbo} & Brazilian coastal ecosystems & 2 & 1 & 15/05/24 \\
         & Brazilian Coastal ecosystems & 1 & 0 & 15/05/24 \\
         & coastal ecosystems & 5 & 4 & 15/05/24 \\
         & Coastal ecosystems & 18 & 18 & 15/05/24 \\
         & Coastal ecosystems & 20 & 22 & 15/05/24 \\
         & ecossistemas costeiros & 36 & 49 & 15/05/24 \\
         & Ecossistemas costeiros & 24 & 33 & 15/05/24 \\
         & ecossistemas costeiros brasileiros & 33 & 35 & 15/05/24 \\
         & Ecossistemas costeiros brasileiros & 12 & 14 & 15/05/24 \\
         & PT-BR-out - ecossistemas costeiros & 18 & 22 & 16/05/24 \\
         & PT-BR-out - Ecossistemas costeiros & 23 & 24 & 16/05/24 \\
         & PT-BR-out - ecossistemas costeiros brasileiros & 26 & 35 & 16/05/24 \\
         & PT-BR-out - Ecossistemas costeiros brasileiros & 10 & 12 & 16/05/24 \\
        \cline{1-5}
        \multirow[t]{13}{*}{4} & Brazilian coastal ecosystems & 19 & 19 & 20/08/24 \\
         & Brazilian Coastal ecosystems & 51 & 59 & 14/09/24 \\
         & coastal ecosystems & 86 & 145 & 16/05/24 \\
         & Coastal ecosystems & 62 & 72 & 17/05/24 \\
         & Coastal ecosystems & 94 & 133 & 17/05/24 \\
         & ecossistemas costeiros & 77 & 106 & 20/08/24 \\
         & Ecossistemas costeiros & 99 & 141 & 06/09/24 \\
         & ecossistemas costeiros brasileiros & 33 & 40 & 20/08/24 \\
         & Ecossistemas costeiros brasileiros & 51 & 59 & 14/09/24 \\
         & PT-BR-out - ecossistemas costeiros & 3 & 2 & 05/09/24 \\
         & PT-BR-out - Ecossistemas costeiros & 11 & 10 & 05/09/24 \\
         & PT-BR-out - ecossistemas costeiros brasileiros & 1 & 0 & 05/09/24 \\
         & PT-BR-out - Ecossistemas costeiros brasileiros & 1 & 0 & 05/09/24 \\
        \cline{1-5}
        \bottomrule
        \end{tabular}
    }
    \caption{Summary of the main experiments}
    \label{tab:gen_results_by_model} 
\end{table*}

Table \ref{tab:gen_results_by_model} summarizes the 26 executions of this stage, divided by the model that was used. The column C0 corresponds to the input passed as initial concept in each run of the algorithm, and the next two columns show how many concepts and subsumptions appear in the output, respectively. The last column is the register of the date each run occurred.
The entries listed with the prefix "PT-BR-out" correspond to the executions using the PT-out code version. The input "Coastal ecosystems" was tested twice per model to be considered as a parameter of each model's normal variation.

\begin{figure}[ht]
    \centering    
    \includegraphics[width=0.45\textwidth]{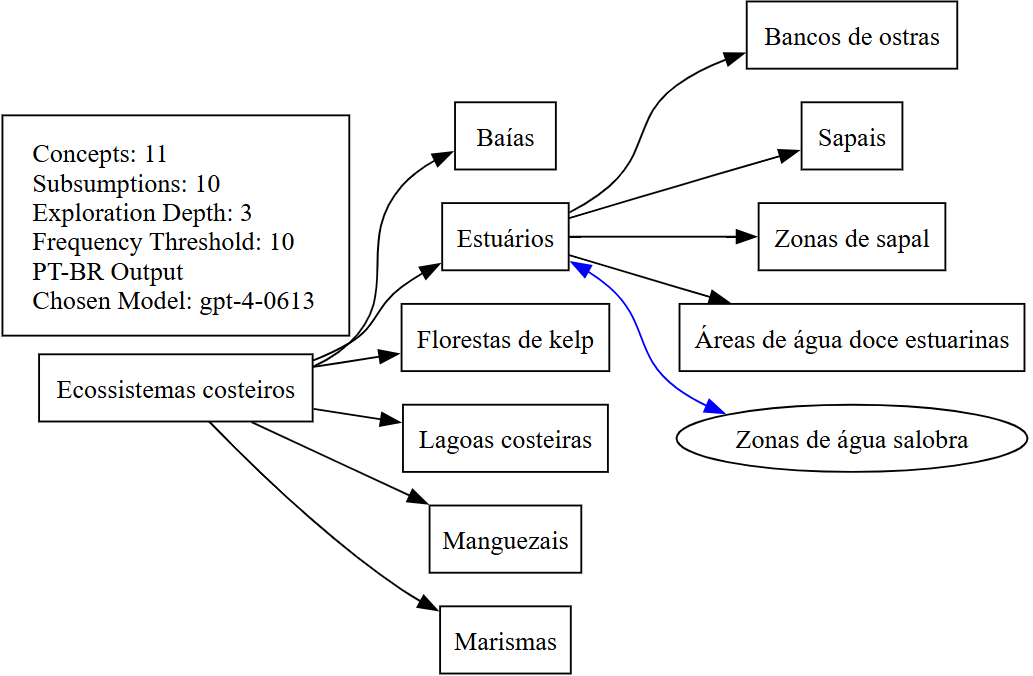}
    \caption{Graphic output generated by passing "Ecossistemas costeiros" as \textit{C0}, 3 as Exploration Depth and 10 as Frequency Threshold to the PT-out code version of the algorithm connected to GPT-4 API}
    \label{fig:graph_out_example}
\end{figure}

The outputs of an execution include different formats of representation for the resulting ontology, such as an OWL file, with all concepts and their definitions, and an SVG file, that allows for the visualization of the produced hierarchy. Figure \ref{fig:graph_out_example} is an examples of the latter.
In the image, the first box from top to bottom on the left summarizes the respective execution, informing how many concepts were added to the built ontology and how many subsumptions connecting them there are. It also shows the values used for the hyperparameters Exploration Depth and Frequency Threshold, and whether this output was generated by the PT-out code version, in which case this box will contain the headline "PT-BR Output" (as is the case here).
The rest of the rectangular boxes each show concepts placed on the ontology, with the arrows going from a concept to its subconcept. At the root is the \textit{C0}, entered as input. The oval shape connected to a concept by a bidirectional blue arrow shows how concepts considered synonyms by the model are represented.

\subsection{Evaluation}

All ontologies produced on the main experiments that turned out with at least 10 concepts were considered valid to be evaluated by the team of domain specialists. As it can be deduced from table \ref{tab:gen_results_by_model}, this means each model had 10 of its generated ontologies formally assessed.

The evaluation occurred through questionnaires, formulated mostly based on the works of Lozano-Tello and Gómez-Pérez \shortcite{2004_Lozano-Tello_Gomez-Perez} — focused on creating metrics that would allow an objective scaling of the adequacy of an ontology to the desired goals — and of Almeida \shortcite{2009_Almeida} — geared towards a more qualitative evaluation of an ontology —. 
Lozano-Tello and Gómez-Pérez \shortcite{2004_Lozano-Tello_Gomez-Perez} motivated the idea of grouping the topics of interest hierarchically, and the preparation for the calculation of a numeric score for each ontology using a five graduations scale. Some of the factors they suggest for the content evaluation also served as inspiration.
From Almeida \shortcite{2009_Almeida}, the main influences were their criteria related to information quality, and the insight to formulate questions focused on assessing to what degree an ontology is succeeding in the goal of modeling the real-world domain and how well the knowledge is being transmitted.

The aspects that were assessed by the specialists are listed below. The ones on the higher-level we call the dimensions. Some of these were defined through a set of factors, that would then be aggregated to compose the score for that dimension. The aggregation of the scores for all the dimensions of an ontology determines their total score. Note that we often refer to a concept and its subconcept as a "parent" and a "child" in relation to each other, and that subconcepts might be connect to more than one parent.
\begin{itemize}
    \item Accuracy
       \begin{itemize}
            \item quality of the definition
            \item appropriateness of the chosen name
            \item how related the concept is to its parent(s)
       \end{itemize}
    \item Relevance — how much this subconcept adds knowledge by being positioned as a child of its parent(s)
    \item Coverage — whether the children of this concept are sufficient to explain it
    \item Precision
       \begin{itemize}
            \item positioning in the ontology 
            \item if the concept is indeed a subconcept (or if is closer to being a part, instance, or something else) of its parent(s)
            \item how precise are its connections (Both missing and unnecessary connections must be penalized)
            \item adequacy of the synonym (if any)
       \end{itemize}    
    \item Information Design
       \begin{itemize}
            \item depth (Number of levels)
            \item size (Number of concepts)
            \item overall evaluation (General grade for the ontology)
       \end{itemize}
\end{itemize}

All listed aspects, except for the ones under the dimension Information Design, needed to be analyzed at the concept level, considering each of their connections. Since, to the best of our knowledge, it was the first in-depth evaluation of the algorithm, it was considered important to go into this level of detail to have a parameter of its performance. 
The assessed aspects were rated mostly using a scale of 1 to 5, where 1 is the top evaluation, and 5 is the worst. By the time of the formulation of the first questionary template, the choice of using an inverted scale seemed more adequate so that all questions would match more closely the logic of the assessment of the missing connections in the output (where fewer is considered better), and the pattern was then followed through the whole evaluation phase for coherence. Questions that presented a different scale on the form had the answers rescaled for the analysis\footnote{For further detail of the conversions, refer to the results processing codes on the repository}.
The forms also provided the opportunity for the respondents to express some impressions that could not be completely translated into a quantitative metric. Those were qualitatively analyzed to compose the results.

The team of evaluators was composed of five academics with a background related to the study of the Ocean, from fields such as Geosciences, Environmental Resource Management and Oceanography, most with experience in projects or studies related to Sustainability. The composition was diverse in age, gender, as well as in the educational level, ranging from undergraduate to post-doctor, and including some specialists with experience in lecturing. While answering the questionnaires, they analyzed the constructed hierarchies considering the visual outputs and the definitions the LLMs provided for each concept.
Since the suitability of ontologies also depends on the context of their application, it was stablish that their assessment should be based on what an elementary school student should be taught about the topic defined by the root concept of each given ontology.

Before the actual evaluation start, a test round was conducted, preparing forms for two smaller outputs, which were replied by the specialists after a basic explanation, followed by a feedback session to both clarify their doubts and to collect their insights to improve the evaluation.
After some adjustments, we proceeded with the assessment of the ontologies produced with GPT-3.5, where all ten valid ontologies were analyzed by all five evaluators.

When the executions using GPT-4 were concluded, however, given the size of the outputs produced, adjustments on the assessment dynamic proved to be necessary.
As a result, simplifications were made on the forms, managing the transition to preserve compatibility with the former template and to avoid losing effectiveness on the evaluation. 
Also, each of the ten forms prepared for this round was only assigned to two evaluators, since even after the modifications, most of the them still required more than double the time of the previous ones to be completed. Therefore, each specialist only judged four of the GPT-4 outputs, taking into consideration the size of the assigned ontologies to keep the workload of each expert similar.


\section{Results}

As mentioned in the previous section, all the outputs with 10 concepts or more were evaluated by the domain experts, meaning they were reviewed and received scores on five dimensions, and exactly ten ontologies made with the aid of each of the models achieved the minimal required expansion. No \textit{C0} was completely left out of the evaluations, but some only achieved the minimal expansion required on one of the models. We can observe in table \ref{tab:gen_results_by_model} that these initial concepts that had insufficient expansion on one of the models often had very significant expansions on the other (considering the average expansion of each model, of course), highlighting the difference in the operation pattern of the same algorithm with each model.

\begin{table*}[ht] 
    \resizebox{\textwidth}{!}{
        \begin{tabular}{lrrrrrrr}
        \toprule
         & Accuracy & Coverage & Precision & Relevance & Information design & Total & Direct Evaluation \\
        Model &  &  &  &  &  &  &  \\
        \midrule
        3.5 & 1.784595 & 2.464992 & 1.637863 & 2.044545 & 2.549630 & 2.096325 & 2.848889 \\
        4.0 & 1.566312 & 1.567132 & 1.860130 & 1.870655 & 2.849074 & 1.942660 & 2.822222 \\
        \bottomrule
        \end{tabular}
    }
    \caption{Average scores of the evaluated ontologies, aggregated by model}
    \label{tab:totals_by_model}
\end{table*}

Table \ref{tab:totals_by_model} shows the average scores, aggregated by model, the outputs received for each dimension — Accuracy, Coverage, Precision, Relevance and Information design —, as well as the calculated mean of these metrics (Total). It also includes the column Direct Evaluation, with the average for the general grades the specialists directly attributed to each ontology they reviewed. Following the logic of the scale, a smaller grade means a better rating.

GPT-4 was better evaluated in most dimensions, but GPT-3.5 surpassed it in Precision and in Information Design. In most of them, the difference was not so significant, varying around 0.2 points.
It is more pronounced, though, in the Coverage dimension, amounting to circa 0.9 point. It seems coherent to assume it is related to the smaller expansion we observe in most ontologies made with GPT-3.5. On the other hand, the greater expansion seems to have made outputs of model 4, on average, excessive and less precise.
It is interesting to notice that, for both models, the direct grading of the ontologies was harsher than the assessment of its aspects individually (summarized in Total). Also, the average score of the models in this metric is basically the same.

\begin{figure*}
    \centering
    \includegraphics[width=\linewidth]{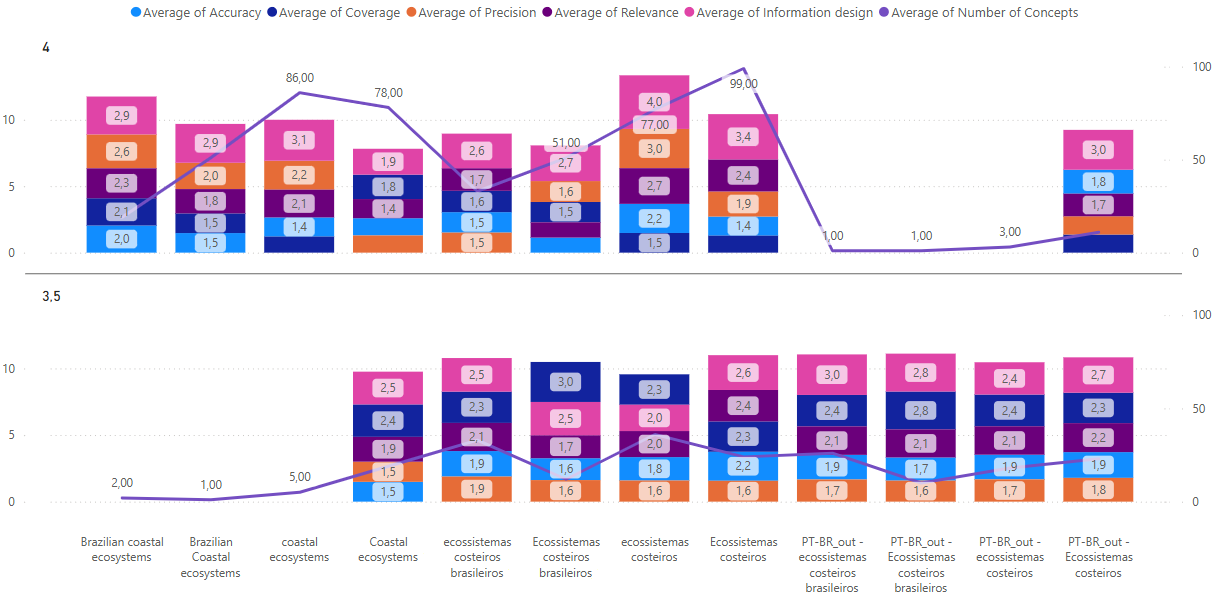}
    \caption{Comparison between models of the averages for the evaluated metrics (bars) and the number of concepts (line) by initial concept}
    \label{fig:Total_fract}
\end{figure*}

\begin{figure}[b]
    \centering
    \includegraphics[width=0.45\textwidth]{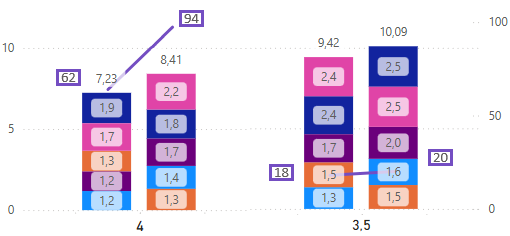}
    \caption{Comparison between repeated executions for \textit{C0} \textit{Coastal ecosystems}. The scheme follows as described in Figure \ref{fig:Total_fract}, with the addition here of the rectangles outlining de number of concepts}
    \label{fig:Total_fract_comp}
\end{figure}

\begin{figure*}
    \centering
    \includegraphics[width=\linewidth]{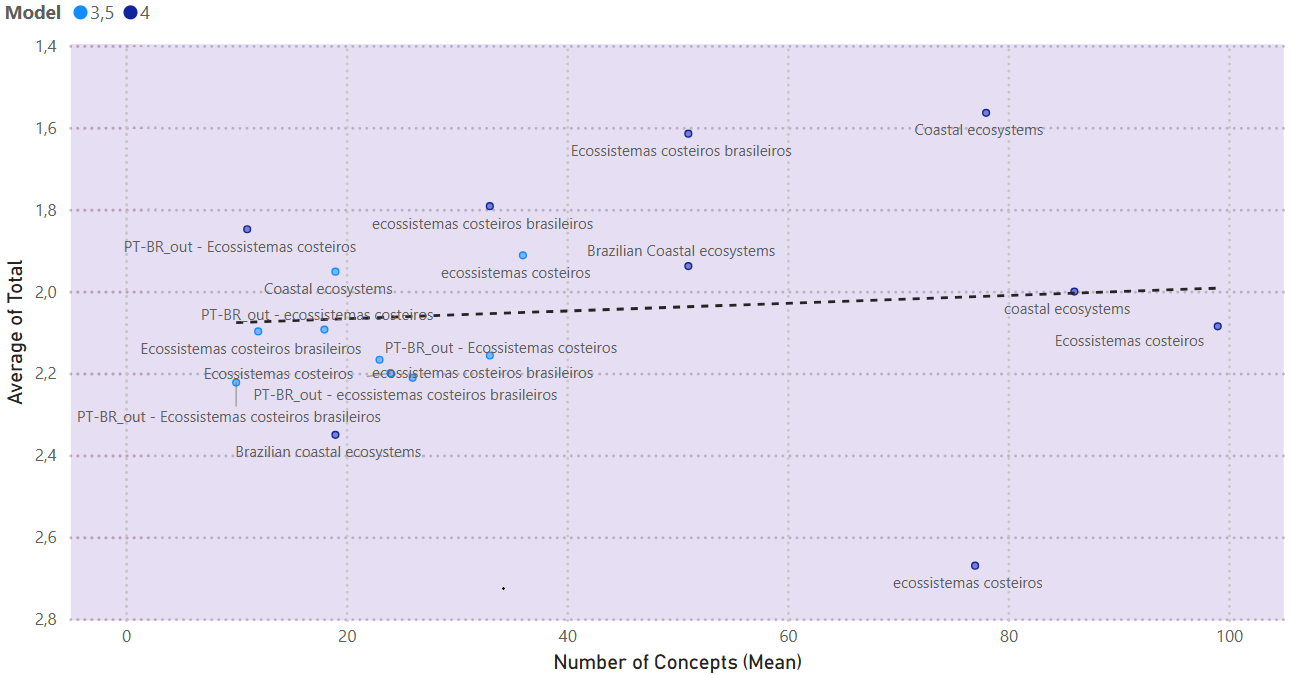}
    \caption{Correlation between number of concepts in the outputs and the average evaluation score (Total) for each initial concept, distinguished by model. The dashed line shows the tendency. For a better visualization, the range of the the y-axis is zoomed and inverted (in accordance with the rating scale).}
    \label{fig:numb_cncptsxtotal}
\end{figure*}

Figure \ref{fig:Total_fract} presents a parallel between the means of these metrics aggregated by \textit{C0} for the two models. The line going through the bars shows the number of concepts for the correspondent output. Here, as in Figure \ref{fig:numb_cncptsxtotal}, the results for the repeated executions of \textit{Coastal ecosystems} are aggregated, but Figure \ref{fig:Total_fract_comp} reveals the individual details and allows for a comparison. The total size of the bars translates to the Total score without the regularization of the average calculation.
Model 3.5 was quite consistent in the ranking of the dimensions, maintaining the order for most cases, with few inversions between adjacent ranks. That is, from best (smallest bars) to worst: Precision; Accuracy; Relevance; Coverage; and Information Design.

For GPT-4, there was greater variation overall. But we can highlight some patterns, such as that all \textit{C0}s had Information Design as the worst score (or one of them, in cases where there was a tie), just like most GPT-3.5 cases. Also, for the most part, Precision is the second worst — and these were, in fact, the two dimensions in which model 3.5 overall surpassed 4 —. It is also possible to observe other similarities to GPT-3 ranks: Accuracy is among the best metrics for most — with the exception of the only output generated with PT-BR-out code version that satisfied the thresholds for assessment in GPT-4, which deviated from the standards in general —; and Relevance is in the central position, or adjacent to it, for all initial concepts.

Comparing the \textit{C0}s that met the evaluation criteria for both models, we can see that model 4 outperformed model 3.5 in all Totals, and in practically all metric comparisons (apart from Information Design). The exception was \textit{ecossistemas costeiros}, which was basically an outlier for GPT-4. It is also noticeable that the number of concepts of model 4's outputs exceeds model 3.5 equivalent's in almost all cases but \textit{ecossistemas costeiros brasileiros}, where both models had the same output size, and in those of PT-BR\_out kind. This demonstrates that size was not determinant for the scoring, since, even in these cases, the metrics favored model 4.

Analyzing the repeated executions separately in Figure \ref{fig:Total_fract_comp}, we observe GPT-3.5 ones were very close in number of concepts, and the average scores only diverge modestly. In model 4, the metrics are more distinct. The second execution generated about 50\% more concepts than the first, and the variation on the ratings is more evident. For both models, though, coincidentally or not, the first execution was better evaluated. We also notice that GPT-4 was considered to perform better overall than GPT-3.5 in all this \textit{C0}'s executions.

Figure \ref{fig:numb_cncptsxtotal} presents the relation of the average Total score and the number of concepts for each \textit{C0} in each model. 
When we look at model 3.5 alone, the tendency for the Total seems to be to improve as the number of concepts increases. Model 4 data mostly corroborates to this tendency, but an inversion in it past a certain output size is outlined, hinting at a limit to the benefit of the ontologies continued expansion.

Looking again at Figure \ref{fig:Total_fract} we may see a reflection of this tendency. With the exception of \textit{Coastal ecosystems}, expansions beyond 70 concepts resulted in scores above 3.0 for Information Design. In addition to the Direct Evaluation, this dimension is composed by the aspects Depth and Quantity. While the GPT-3.5's outputs were mostly rated in those aspects respectively as "sufficient" (= 1) and "moderately insufficient"(= 2.5), GPT-4's were mainly judged as "quite exaggerated"(= 4.5) in both.

Analyzing the effect the variations on the inputs had on the outputs and the corresponding evaluations, we observe that capitalization on GPT-4 produced larger ontologies, on average, with consistently better evaluations. For GPT-3.5, although, in general, capitalized \textit{C0}s also had slightly better expansions, there is almost no difference in the grades.

The language factor seemed to have a greater impact on GPT-3.5. Almost all \textit{C0}s that achieved the 10 concept threshold for evaluation were in Portuguese. The only initial concept in English to obtain significant expansion was \textit{Coastal ecosystems}, with a median number of concepts regarding the results for the model.
The expansion of Model 4 is less constant in terms of language. The set that was evaluated is composed of half of each type, but the grades were better for \textit{C0}s in English.
However, its performance in terms of expansion is much worse when there is an explicit request for the output to be in PT-BR. The two models seem to expand less in the PT-BR\_out code version, in fact, but GPT-4 suffers a greater impact.

As for the reference to Brazil in the \textit{C0}s, the concepts were further expanded by model 4 without it. On GPT-3.5, this also occurred, but the gap is narrow, and the results are actually a blend. Regarding the scores, the difference between the Totals is greater in model 3.5.
The impact of this \textit{C0} variation was also examined regarding the mixing of languages in the outputs, which sometimes occurred for initial concepts in Portuguese.
For GPT-4, in all executions with initial concepts in PT-BR that had relevant expansion, there was some level of English mixed with Portuguese. The exception was the one generated with the PT-BR\_out version of prompts.
As expected, in model 3.5, the ones with the language coercion also did not have any mixing. On the ones without the restriction, the mixing was less frequent than for GPT-4, with only two outputs exhibiting that behavior.

Studying the general comments the evaluators left for the ontologies, the theme of exaggerated expansions for the outputs of model 4 is recurrent. Many reported unnecessary or repetitive concepts within the ontology (usually in the form of multiple inclusions not marked as synonyms of the same concept, with minimal variation. For example, adding to the hierarchy both its plural and single form, or adding versions carrying specific adjectives that do not bring relevant new information). There are also numerous mentions of incorrect or confusing groupings and classifications, and even the invention of terms. Some also commented on finding some definitions limited to a restricted context (e.g., only covering physical aspects and not approaching the characteristics of life in the ecosystem). 

For the model 3.5, some of the same issues appeared on the feedbacks, such as the invention of terms, usually naming incorrectly a concept that does exist; problems in the organization of the hierarchy; lacking descriptions; and some unnecessary concepts or repetition unmarked as synonyms as well — although criticism on this topic was less recurrent and emphatic than for GPT-4. In fact, it was more frequent the contrary complaint, recommending the inclusion of terms or a greater expansion in specific points —. Missing links were also mentioned, as well as incorrect generalizations and inconsistency in granularity.

After all the evaluation forms had been completed, a voluntary form was passed to the specialists requesting they shared their thoughts on the concept of ontologies, the results, and their participation in this research.
The evaluation of their experience turned out positive, with highlights for the opportunity to reflect on the use of ontologies and on the conceptual structure of the theme, considering that a good mental exercise. It has been mentioned, however, that the format the evaluation was conducted, with the long and detailed questionnaires, proved cumbersome at times.
Most of the respondents declared that, previous to their engagement on the project, they had little or no familiarity with ontologies. After the experiment, their impression was that this form of organization of knowledge is useful. They see application for it, especially for educational purposes, facilitating the exposition of knowledge in a didactic way.

All feedbacks emphasized the necessity of refinement on the method. They reported problems with the inconsistent level of granularity, with some of the outputted ontologies (or parts of it) being too superficial, while others were too detailed. It was observed that the deeper the level, the messier the information would appear, demanding more adjustments. Still, the majority of the respondents considered the outputs from both models generally satisfactory, and claimed to have found the automation of the construction of the ontologies useful. It wasn’t unanimous, though. In one case, it was pointed out that, due to the relatively high occurrence of incoherencies, the manual method would be preferable.

\section{Discussion} \label{sec:discussion}

Examining the outputs and their assessment, we can easily notice that GPT-4 produced, on average, larger ontologies than GPT-3.5, and that most metrics were more favorable to the newer and larger model. However, the score advantage is narrow and not consensus, meaning that the considerable difference between the costs is not directly translated to results. Also, the extra expansion included a high level of redundancy and hallucinations. We must observe, however, that the base algorithm had been mostly tested on GPT-3 during its design in the original study, and no variation in language or format for the input had been consistently explored. Possibly the prompts would benefit from some refinement to manage different kinds of input and to better restrain concept expansion for a verbose model, such as the version of GPT-4 tested by the present work.

The test with the repeated \textit{C0}s, conducted minutes away from each other for each model, serves as a thermometer for the variability within the models. Analyzing the outputs directly, we observe that, even in the case of model 3.5, where the metrics turned out very similar, the structures have distinctions. But the divergence on model 4 was certainly more pronounced. This kind of inconsistency in a system might lead to distrust. Of course, the temperature of the models can be adjusted as means to reduce this effect. On the other hand, doing so will diminish the “creative” effect, which is desirable in many applications where LLMs are being employed, as is usually the case for concepts and relations suggestions for an ontology. Thus, finding the balance between “spontaneous” and reliable is another challenge.

The variation in the performance of each model due to linguistic aspects brought some surprises. It was expected that the context approached by each kind could exhibit differences, but we believed the average of concepts would be similar, if not higher in the main training language of the models. 
Even considering our theory that content about the Brazilian coast on the internet was more abundant in Portuguese, we did not anticipate such difficulty for GPT-3.5 on the expansion of most of the concepts in English, which was not shared by GPT-4.

The struggle of the newer model with the coercion for the outputs to be in PT-BR was even more unforeseen, especially considering how GPT-4 expanded well the concepts in Portuguese too without the restriction. It is true that these outputs mixed the languages, but it remains unclear why the explicit request for Portuguese basically extinguished the expansion of the concepts for this model, and why the effect of the variations was so different on each model. It is convenient to mention that the PT-out executions for model 4 were each tested more than once, even on different days, to confirm it was not a bug. The results section only presented the data of the original executions since the retries generated the same outcome.

Although we could observe a trend of declining quality when outputs expand excessively, more tests would be required to estimate an optimal number of concepts, as well as to draw conclusions about other aspects.
Besides, human evaluation surely is sensitive to differences in the interpretation of the participants due to personal experiences, preferences and knowledge, among other variables beyond our control. We understand that the number evaluators employed in this research is also not enough for definitive verdicts. However, this kind of evaluation is quite costly and we did not have the conditions to extend the assessments.

Despite some reported confusions, inventions, and problems with the granularity, we consider the models were both successful in extracting concepts pertinent to the domain and arranging the hierarchical relations mostly with coherence. The merit is also majorly on the algorithm constructed and the engineering of the prompts conducted by Funk et al~\shortcite{2023_Onto_LLMs}, not simply on the use of LLMs. The model is the judge, but the systematic search for new terms, the verification process design, and the traversal algorithm for inserting the terms in the graph under construction conducted the models through the steps to complete the task.

Still, we judge that none of the models showed the capacity to construct entirely satisfactory ontologies autonomously with this technique. As highlighted, even GPT-4, which achieved the best grades overall, received a great deal of criticism on its outputs, and presented a fair amount of hallucinations and inconsistency over the executions. This understanding is, in fact, in accordance with the conclusions and recommendations in the reviewed literature. \cite{2024_Du-etal} states that “while the idea of fully automatic ontology construction is appealing, especially for handling large volumes of data or complex domains, it is worth mentioning that full automatic construction for ontology by a system is still a significant challenge and it is not likely to be possible”. And the recommendation to integrate “human-in-the-loop approaches with expert involvement [to] enhance ontology relevance and accuracy” \cite{2023_Giglou-etal} is also recurrent, even appearing in the work of Funk et al~\shortcite{2023_Onto_LLMs}, the base for these experiments.

\section{Conclusions and Future Work}

This project could successfully conduct the experiments based on the method developed by Funk et al~\shortcite{2023_Onto_LLMs} to study the construction of ontologies in the context of the Brazilian Maritime Coast. It was not possible to arrive at a fully satisfactory hierarchical conceptualization to organize the knowledge about the Blue Amazon with the employed methodology. One of the reasons was the tests showed it was convenient to choose a more restrict initial concept than initially planned in order to achieve more substantial outputs for the assessment. The choice of \textit{Coastal ecosystems}, with the variation on the format of the input, produced diverse results to analyze the effects on the algorithms.

Another highlight was the possibility of repeating the experiments with two models of the same family with different sizes, and very different costs, to compare their performance. Additionally, the use of human specialists on the domain allowed for broader examination of the outputs.
The evaluation revealed model GPT-4 performs slightly better overall. Yet, it struggles with drawbacks such as hallucinations and redundancy. Plus, in both models there was inconstancy on the level of granularity of the conceptualization and incoherencies on some relations. Thus, the produced ontologies should not be directly utilized without a revision. They may serve, though, as drafts that could be refined and expanded to produce adequate symbolic representation of the domain.

The results serve to corroborate with the predominant view presented on the researched literature. Although LLMs are suitable tools to aid on the development of ontologies, at this point, even advanced models do not demonstrate to be safe to produce functional ontologies on their own. Beyond the technical challenges of the technology, the process of constructing an ontology is entangled with specific motivations. The complete context of the desirable applications, so far, seems to be too difficult to infuse in AI technology.

Thus, we conclude the method is valid and particularly interesting to aid in the expansion of OL to domains where existing specific ontologies are lacking, especially if applied under the supervision of domain experts.


We believe this research also has great potential for expansion beyond the scope of this project. As suggested by the article itself of Funk et al~\shortcite{2023_Onto_LLMs}, as a relevant direction for future work, it is also possible to experiment with the construction of ontologies that are more expressive, adding other kinds of relations and, possibly, even rules, as disjointness, for example.
Another interesting suggestion they make is fine-tuning for domain-specific ontology construction, training the model with curated information about the intended subjects.

Another line of work that could be of great value is the modification to the prompts to include the human on the loop approach. Note that, as described by Noy and Mcguinness~\shortcite{Onto_101}, ontology development is necessarily an iterative process. After an initial version of the ontology is defined, it can be evaluated and debugged by using it in applications or problem-solving methods, or by discussing it with experts in the field, or both. They state that, as a result, the original design will almost certainly need to be revised. Therefore, turning this technique into a practical instrument for specialists to work for ontology development seems like a natural idea.

\section{Acknowledgments}

This work was carried out at the Center for Artificial Intelligence (C4AI-USP), with the support of the University of São Paulo, the São Paulo Research Foundation (FAPESP, grant \#2019/07665-4) and by the IBM Corporation. Vivian Magri A. Soares was also supported by CAPES.

\bibliographystyle{kr}
\bibliography{kr-sample}

@article{2023_Onto_LLMs,
  author       = {Maurice Funk and
                  Simon Hosemann and
                  Jean Christoph Jung and
                  Carsten Lutz},
  title        = {Towards Ontology Construction with Language Models},
  journal      = {CoRR},
  volume       = {abs/2309.09898},
  year         = {2023},
  url          = {https://doi.org/10.48550/arXiv.2309.09898},
  doi          = {10.48550/ARXIV.2309.09898},
  eprinttype    = {arXiv},
  eprint       = {2309.09898},
  bibsource    = {dblp computer science bibliography, https://dblp.org}
}

@article{2023_SCRE,
  author       = {Vahideh Reshadat and
                  Alp Akcay and
                  Kalliopi Zervanou and
                  Yingqian Zhang and
                  Eelco de Jong},
  title        = {{SCRE:} special cargo relation extraction using representation learning},
  journal      = {Neural Comput. Appl.},
  volume       = {35},
  number       = {25},
  pages        = {18783--18801},
  year         = {2023},
  url          = {https://doi.org/10.1007/s00521-023-08704-9},
  doi          = {10.1007/S00521-023-08704-9},
  bibsource    = {dblp computer science bibliography, https://dblp.org}
}

@article{2022_BLAB,
  author       = {Paulo Pirozelli and
                  Ais B. R. Castro and
                  Ana Luiza C. de Oliveira and
                  Andr{\'{e}} Seidel Oliveira and
                  Fl{\'{a}}vio Nakasato Ca{\c{c}}{\~{a}}o and
                  Igor Cataneo Silveira and
                  Jo{\~{a}}o G. M. Campos and
                  Laura C. Motheo and
                  Leticia F. Figueiredo and
                  Lucas Francisco Amaral Orosco Pellicer and
                  Marcelo A. Jos{\'{e}} and
                  Marcos Menon Jos{\'{e}} and
                  Pedro de M. Ligabue and
                  Ricardo S. Grava and
                  Rodrigo M. Tavares and
                  Vin{\'{\i}}cius B. Matos and
                  Yan V. Sym and
                  Anna H. R. Costa and
                  Anarosa A. F. Brand{\~{a}}o and
                  Denis Deratani Mau{\'{a}} and
                  F{\'{a}}bio Gagliardi Cozman and
                  Sarajane M. Peres},
  title        = {The BLue Amazon Brain {(BLAB):} {A} Modular Architecture of Services
                  about the Brazilian Maritime Territory},
  journal      = {CoRR},
  volume       = {abs/2209.07928},
  year         = {2022},
  url          = {https://doi.org/10.48550/arXiv.2209.07928},
  doi          = {10.48550/ARXIV.2209.07928},
  eprinttype    = {arXiv},
  eprint       = {2209.07928},
  bibsource    = {dblp computer science bibliography, https://dblp.org}
}

@misc{Code_Llama,
      title={Code {L}lama: Open Foundation Models for Code}, 
      author={Baptiste Rozière and Jonas Gehring and Fabian Gloeckle and Sten Sootla and Itai Gat and Xiaoqing Ellen Tan and Yossi Adi and Jingyu Liu and Tal Remez and Jérémy Rapin and Artyom Kozhevnikov and Ivan Evtimov and Joanna Bitton and Manish Bhatt and Cristian Canton Ferrer and Aaron Grattafiori and Wenhan Xiong and Alexandre Défossez and Jade Copet and Faisal Azhar and Hugo Touvron and Louis Martin and Nicolas Usunier and Thomas Scialom and Gabriel Synnaeve},
      year={2023},
      eprint={2308.12950},
      archivePrefix={arXiv},
      primaryClass={cs.CL}
}

@article{Srv2018,
  author       = {Muhammad Nabeel Asim and
                  Muhammad Wasim and
                  Muhammad Usman Ghani Khan and
                  Waqar Mahmood and
                  Hafiza Mahnoor Abbasi},
  title        = {A survey of ontology learning techniques and applications},
  journal      = {Database J. Biol. Databases Curation},
  volume       = {2018},
  pages        = {bay101},
  year         = {2018},
  url          = {https://doi.org/10.1093/database/bay101},
  doi          = {10.1093/DATABASE/BAY101},
  bibsource    = {dblp computer science bibliography, https://dblp.org}
}

@article{Onto_101,
author = {Noy, Natasha and Mcguinness, Deborah},
year = {2001},
month = {01},
pages = {},
title = {Ontology Development 101: A Guide to Creating Your First Ontology},
volume = {32},
journal = {Knowledge Systems Laboratory}
}

@article{2004_Lozano-Tello_Gomez-Perez,
  author       = {Lozano-Tello, Adolfo and G{\'o}mez-P{\'e}rez, Asunci{\'o}n},
  title        = {{ONTOMETRIC:} {A} Method to Choose the Appropriate Ontology},
  journal      = {J. Database Manag.},
  volume       = {15},
  number       = {2},
  pages        = {1--18},
  year         = {2004},
  url          = {https://doi.org/10.4018/jdm.2004040101},
  doi          = {10.4018/JDM.2004040101},
  bibsource    = {dblp computer science bibliography, https://dblp.org}
}

@article{2009_Almeida,
  author       = {Mauricio Barcellos Almeida},
  title        = {A proposal to evaluate ontology content},
  journal      = {Appl. Ontology},
  volume       = {4},
  number       = {3-4},
  pages        = {245--265},
  year         = {2009},
  url          = {https://doi.org/10.3233/AO-2009-0070},
  doi          = {10.3233/AO-2009-0070},
  bibsource    = {dblp computer science bibliography, https://dblp.org}
}

@article{2025_Amalki-etal,
  title={Deep Learning-Driven Ontology Learning: A Systematic Mapping Study},
  author={Amalki, Asma and Tatane, Khalid and Bouzit, Ali},
  journal={Engineering, Technology \& Applied Science Research},
  volume={15},
  number={1},
  pages={20085--20094},
  year={2025}
}

@misc{2024_Perera-Liu,
  title={Exploring large language models for ontology learning},
  author={Perera, Olga and Liu, Jun},
  year={2024}
}

@misc{2024_Du-etal,
      title={A Short Review for Ontology Learning: Stride to Large Language Models Trend}, 
      author={Rick Du and Huilong An and Keyu Wang and Weidong Liu},
      year={2024},
      eprint={2404.14991},
      archivePrefix={arXiv},
      primaryClass={cs.IR},
      url={https://arxiv.org/abs/2404.14991}, 
}

@misc{2023_Giglou-etal,
      title={LLMs4OL: Large Language Models for Ontology Learning}, 
      author={Hamed Babaei Giglou and Jennifer D'Souza and Sören Auer},
      year={2023},
      eprint={2307.16648},
      archivePrefix={arXiv},
      primaryClass={cs.AI},
      url={https://arxiv.org/abs/2307.16648}, 
}

\end{document}